\documentclass[runningheads]{llncs}

\usepackage{cite}
\usepackage{times}
\usepackage{amsmath,amssymb,amsfonts}
\usepackage{algorithm,algorithmic}
\usepackage{subfigure}
\usepackage[pdftex]{graphicx}
\usepackage{textcomp}
\usepackage{color}
\usepackage{xcolor}
\usepackage{pgfplots}
\pgfplotsset{compat=1.16}
\usepackage{tikz}
\usepackage{bm}
\usepackage[pagebackref=true]{hyperref} 
\usepackage{soulutf8}
\usepackage{booktabs}
\usepackage{multirow}
\usepackage{amsmath}
\usepackage{amsfonts,bm}
\usepackage{graphicx,lipsum}
\usepackage{todonotes}
\usepackage{floatpag}
\usepackage{xspace}

\newcommand{\toolname}{\textsc{AGNES}\xspace}
\newcommand{\abstractmethod}{AbsSM\xspace}
\newcommand{\approxmethod}{AproxSM\xspace}

\begin{document}
\title{\toolname: Abstraction-guided Framework for Deep Neural Networks Security\thanks{This research was funded in part by
    the EU under project 864075 CAESAR, the project Audi Verifiable AI, and the BMWi funded KARLI project (grant 19A21031C).}}
\author{Akshay Dhonthi\inst{1,2} \and
Marcello Eiermann\inst{3} \and
Ernst Moritz Hahn\inst{2} \and
Vahid Hashemi\inst{1}}
\authorrunning{A. Dhonthi et al.}
\institute{AUDI AG, Auto-Union-Stra\ss e 1, 85057, Ingolstadt, Germany \and
Formal Methods and Tools, University of Twente, Enschede, Netherlands \and
Information and Computing Sciences, Utrecht University, Utrecht, Netherlands\\}
\maketitle              % typeset the header of the contribution
\begin{abstract}
Deep Neural Networks (DNNs) are becoming widespread, particularly in safety-critical areas.
One prominent application is image recognition in autonomous driving, where the correct classification of objects, such as traffic signs, is essential for safe driving.
Unfortunately, DNNs are prone to \emph{backdoors}, meaning that they concentrate on attributes of the image that should be irrelevant for their correct classification.
Backdoors are integrated into a DNN during training, either with malicious intent (such as a manipulated training process, because of which a yellow sticker always leads to a traffic sign being recognised as a stop sign) or unintentional (such as a rural background leading to any traffic sign being recognised as ``animal crossing'', because of biased training data).
  
In this paper, we introduce \toolname, a tool to detect backdoors in DNNs for image recognition.
We discuss the principle approach on which \toolname is based.
Afterwards, we show that our tool performs better than many state-of-the-art methods for multiple relevant case studies.
  
%The abstract should briefly summarize the contents of the paper in 15--250 words.
\end{abstract}
\section{Introduction}
% Motivation
\emph{Deep Neural Networks} (DNNs) are widely used nowadays in safety-critical application such as automotive~\cite{faster2015towards, chen2015deepdriving}, avionics~\cite{dmitriev2021toward} and medical~\cite{chen2022recent} industries. In particular, in the automative domain, the safety-critical applications range from automated driver assistance systems (ADAS) such as Automated Emergency Braking Systems (AEBS) up to fully Automated Driving (AD) vehicles. 
In such applications, machines take over more and more critical decisions.
This leads to two types of risks: \emph{safety} and \emph{security}~\cite{salay2017analysis}.
The former describes functional safety, i.e., protecting the environment and other road users from the machines. 
In contrast, the latter describes the protection of machines from the environment and other external interventions.
In this work, we describe the tool \toolname, which focuses on security risks in DNNs.
We specifically focus on classification functions where, given an image, the DNN predicts to what class the image belongs.
This is an important task particularly in safety-critical domains such as the traffic sign detection function in an autonomous system. 

A security risk occurs when a DNN does not predict the true class of an image, but instead focuses on regions in the image which are not intended to influence the classification.
Such behaviour results if the dataset used to train a DNN is intentionally manipulated by adding triggers to a small part of the dataset or when the DNN learns from regions in the training images that are irrelevant for decision-making.
We call such behaviours \emph{backdoors} of the network~\cite{gu2017badnets, liu2018trojaning}.
Further, with \emph{trigger}, we denote the part of the image which activates the backdoor.
A trigger may be as simple as a small patch in a specific location of the image, or it can also be a nearly invisible watermark spread over the whole image~\cite{gu2017badnets}.
It could also be very complex: with inappropriate training, a DNN might focus on the background of an image to guess the traffic sign to be detected rather than on the traffic sign itself\cite{saha2020hidden, zhao2020clean}.
Additionally, proving the presence of a backdoor, describing its triggers, and removing backdoors are highly complex tasks.
This is particularly true if one is unaware of the exact type of possible triggers.

Identification of backdoors has been an interest for a while now, and several techniques have been developed in recent years~\cite{rauker2023toward, huang2020survey}.
Backdoor detection can be performed by synthesising the trigger and then using it to analyse the performance of DNNs.
Apart from techniques that synthesise triggers, other defence methods directly defend from backdoors either during runtime~\cite{ma2019nic} or post-hoc analysis~\cite{kolouri2020universal, huang2020one}. 
However, in this work, we focus on techniques based on trigger-synthesis because we can visualise the backdoors.
Mainly, the techniques that are worth mentioning are Neural cleanse (NC)~\cite{wang2019neural}, Artificial Brain Stimulation (ABS)~\cite{liu2019abs}, ABS+Exray~\cite{liu2022complex}, and  K-Arm~\cite{shen2021backdoor}.

NC is the first proposed trigger-synthesis based defence method. Here, potential trigger patterns are obtained for each class via trigger optimisation \cite{li_backdoor_2022}. 
The ABS technique is based on stimulating/changing every neuron output in the hidden layers and propagating this change up to the output layer to see if it impacts the network output. 
When the neurons are stimulated, the neurons that change the network's output are considered \emph{compromised neuron candidates} (CNCs) and are later used to generate triggers.
The ABS+Exray method uses a trigger inversion technique to generate triggers and then analyses if the triggers contain features that are not considered natural distinctive features between the victim (the original class) and the target class.
The K-Arm optimization method, which is built on top of ABS, propose a Reinforcement Learning based method to detect backdoors.
We integrate ABS in our tool because it does not require pre-knowledge of the trigger type. This helps to synthesise triggers by utilising only benign images, making it more interesting to automotive applications where the trigger is unknown.

However, as mentioned in~\cite{dhonthi2023backdoor}, the ABS technique is slow because the stimulation of neurons is time-consuming.
ABS analyses all the features in the hidden layers sequentially to identify backdoors.
Doing so is inefficient due to the large number of features in the hidden layers.
Also, ABS generates an intermediate list of neurons called compromised neuron candidates, which contain many false positives.
This means that most of these neurons may not lead to identifying triggers~\cite{shen2021backdoor}.
To address these issues, we propose a way to abstract the inner parameters of a network such that the identification of compromised neurons in the network is fast and robust.
We apply our method over several model architectures, trigger types, and trigger parameters (such as trigger size and transparency) to evaluate its performance.

\begin{figure}[t]
\includegraphics[width=\textwidth]{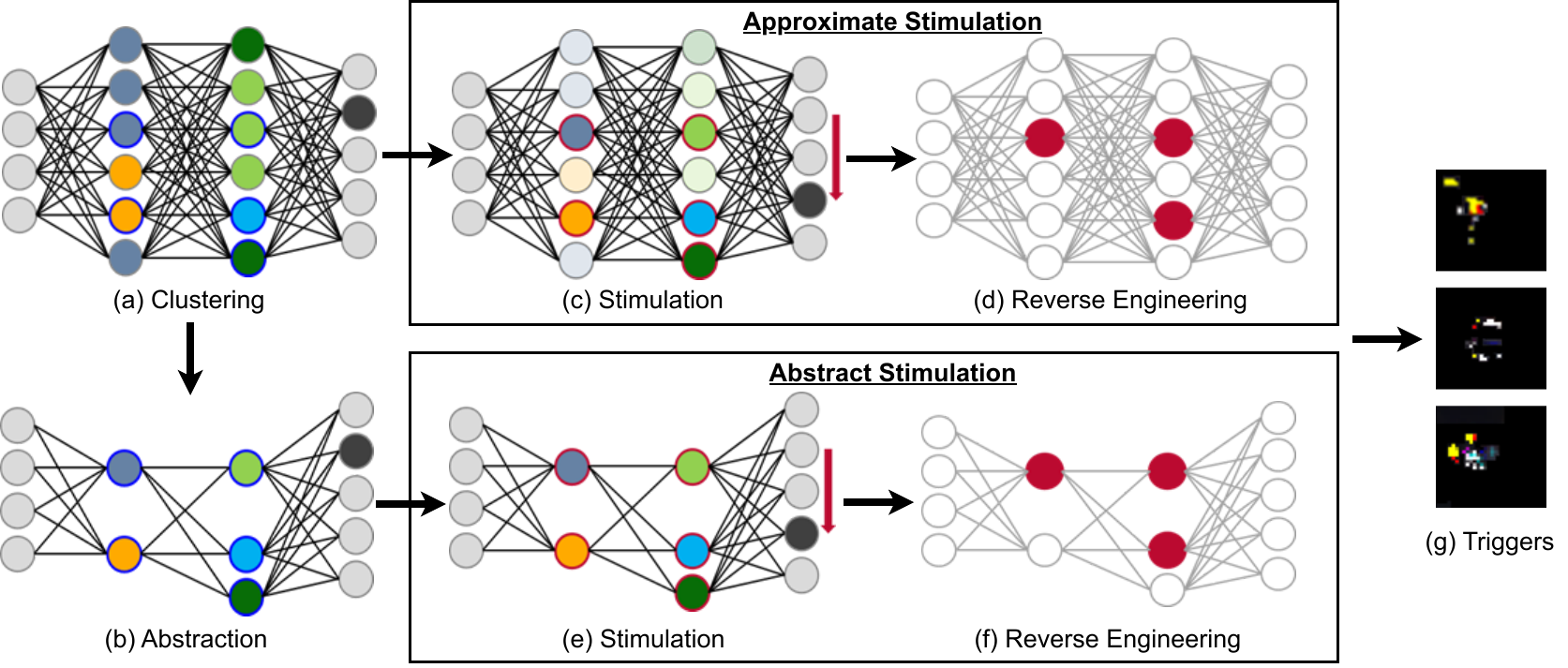}
\caption{\toolname Framework. The colours on neurons represent various clusters. The neurons with the dark blue outline are cluster representatives of each cluster. The neurons with red outlines undergo stimulation while the rest are skipped. The red neurons in the reverse engineering step are the compromised neurons}
\label{Fig: Framework}
\vspace{-0.50cm}
\end{figure}

We depict the principle workflow of our tool \toolname in Fig~\ref{Fig: Framework}.
It can find the most critical neurons quickly and outperforms state-of-the-art (SOTA) techniques in identification of some types of triggers.
Our tool works with the DNN as a black-box in so far as that it only needs the trained network and a small benign dataset, but not the training data or further information about the nature of potential triggers.
\toolname abstracts the neural network by clustering the neurons in each layer and abstracts each cluster to its \emph{cluster representatives} (CR) (neuron that is closest to the cluster centre) based on DeepAbstract~\cite{ashok2020deepabstract} technique.
We propose two methods:
The first one is \emph{Abstract Stimulation Method} (\abstractmethod), where we stimulate the abstract network.
The second one is the \emph{Approximate Stimulation} (\approxmethod), where we cluster each hidden layer based on their activation values and then stimulate only the cluster representatives for analysis.
We evaluate both techniques on several model architectures, trigger types, and trigger parameters such as trigger size and transparency. 
In many cases (cf.~Sec~\ref{Sec: Experiments}), \toolname performs better in terms of time complexity and accuracy than other techniques.

Overall, our contributions in this paper are as follows:
\nolinebreak
\begin{itemize}
    \item We introduce the tool \toolname, which improves SOTA backdoor identification techniques in terms of performance and runtime.
    \item We propose two methods, namely \abstractmethod and \approxmethod for stimulation analysis. The former is a precise method which abstracts the network for analysis while the latter is an approximate method in which instead of abstracting the network, a smaller set of neurons for analysis is extracted. The proper method selection for analysis is done automatically by \toolname.
    \item We perform an extensive evaluation of our tool on SOTA model architectures for traffic sign classification problem while adhering to the latest machine learning frameworks such as TensorFlow $2.0$~\cite{abadi2016tensorflow} and PyTorch~\cite{paszke2019pytorch}.
\end{itemize}
\iffalse

The paper starts with preliminaries in Sec~\ref{Sec: Preliminaries}, which includes a formal definition of DNNs, the abstraction technique, and the stimulation technique~\cite{liu2019abs}.
We then give a detailed description of the methodology in Sec~\ref{Sec: Methodology}.
Finally, our results in Sec~\ref{Sec: Experiments} show the performance of our tool on various well-known DNN architectures and several kinds of triggers.
We conclude the paper in Sec~\ref{Sec: Conclusion} with several future directions.
\fi

\section{Preliminaries} \label{Sec: Preliminaries}
\vspace{-0.30cm}
In this section, we introduce the key components of \toolname.

\vspace{-0.30cm}
\subsection{Deep Neural Networks}

\begin{figure}[t]
  \begin{center}
    \scalebox{0.9}{\includegraphics[width=\textwidth, trim={0.0cm 0.45cm 0.0cm 0.0cm},clip]{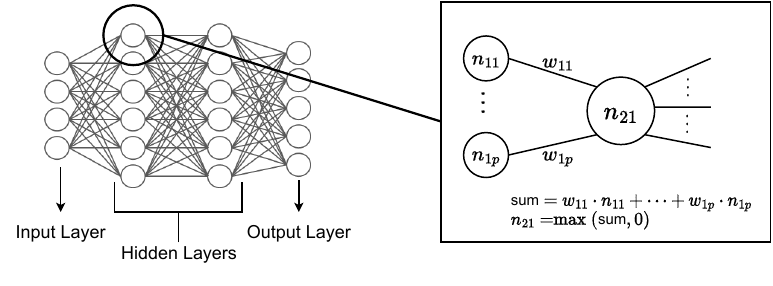}}
  \end{center}
  \vspace{-0.30cm}
  \caption{The left image depicts the DNN structure, and the right image depicts the computation of a single neuron output. We show, as an example, computation for neuron $n_{21}$.
  }
  \label{Fig: DNN}
  \vspace{-0.40cm}
  \end{figure}

In Figure~\ref{Fig: DNN}, we provide an example of a simple DNN.
The left part provides a high-level overview of the network, depicting the connections between the neurons.
The input layer takes the shape of an image, where \emph{shape} represents the total number of elements and dimensions.
The image can have the shape $[l, w, c]$, where l and w are the length and width of the image, and c represent the channels typically 3 (R, G, B).
The output layer is the shape of the number of classes to which an image can be classified.
The hidden layers can be but not limited to convolutional, activation, pooling, or fully connected layers.
The right part details how a single neuron works in a fully connected layer.
The neuron detailed receives input values from its predecessor neurons, weighted with the strength of their connections.
These values are summed up, resulting in a real value.
On this sum, we apply the \emph{activation function} of this neuron, which is usually the same for all neurons of the layer.
In this case, we use \emph{ReLU} (maximum of 0 and the weighted input sum), which is the most commonly used activation function nowadays.

DNNs can approximate arbitrary functions~\cite{goodfellow2016deep} and perform a wide range of tasks.
Their advantage is that they can be trained on a large set of examples to provide the correct output on inputs sufficiently similar to the ones they have been trained with.
Their main disadvantage is that they do not have any easily understandable structure, consisting mostly of a large number of real numbers representing the weights connecting the individual neurons.
This is also why it is easy for them to have hidden backdoors that are hard to find in a DNN~\cite{gu2017badnets}.

\vspace{-0.30cm}

\subsection{Abstraction}
\label{Sec: Abstraction}
For this work, we use DeepAbstract~\cite{ashok2020deepabstract} as the abstraction technique, but any abstraction method would be suitable.
DeepAbstract clusters the neurons in a layer via the k-means function.
For each cluster, we chose a representative: the neuron closest to the centroids of the respective cluster.
This abstraction technique only works for networks with Fully Connected (FC) layers. 
Therefore, we propose an approximate technique, \approxmethod, in Sec~\ref{Sec: Methodology} to handle non-FC networks such as convolutional networks.
Later after identifying clusters, all the neurons in a cluster are dissolved into the respective CR by redirecting all the weights into the CR.
Since the activation values of all the neurons in a cluster are similar, the loss of information flow resulting from summing all the connections in the cluster to the CR will be low.

We obtain the optimal number of clusters by making sure the loss in accuracy of the DNN is minimal after clustering.
The method is illustrated in Figure~\ref{Fig: Framework} (b), where the original network is abstracted using only CR.
Neurons belonging to the same class mostly behave alike.
Therefore, our intuition is that stimulating the CR neurons can alone identify the neurons responsible for triggering backdoors.

\vspace{-0.30cm}
\subsection{Stimulation}
We utilise the stimulation technique from ABS, where each neuron is systematically stimulated to check whether the output of the network changes.
The stimulation is done as follows:
We increase or decrease the neuron output under analysis in a hidden layer by adding \emph{stimulation value}.
Positive/negative stimulation values will increase/decrease the neuron output, respectively.
While doing so, we keep the rest of the neuron outputs in that layer as they are.
A neuron is then marked as a \emph{compromised neuron candidate} (CNC) if, under some stimulation value, the output class of the network will shift when stimulated neruon value is forward propagated to the output layer.

We compute stimulation outputs for images from different classes, and analyse whether the output shift is consistent.
This means that we identify the neurons such that for a specific stimulation value, the network output remains the same regardless of the class of the image.
Such a neuron is deemed compromised as it completely influences the network's output.
Our goal is to identify such neurons with high precision.
ABS performs this task systematically for all the neurons in the hidden layers, which can be slow for large models.
Instead, in \toolname, we restrict the stimulation analysis only to the CR.
Doing so can be efficient with respect to time.
This process is further explained in Sec~\ref{Sec: Methodology}.

\section{Methodology} \label{Sec: Methodology}
We propose two methods: stimulation over the abstract network (\abstractmethod) and over the original network (\approxmethod).
The difference between both techniques concerns the neurons and the network used for stimulation, as depicted in Fig~\ref{Fig: Framework}.
The former technique stimulates all the neurons but works with the abstract network.
However, the latter technique stimulates the CR on the original network and ignores the rest of the neurons for stimulation analysis.
This section will first describe the clustering and abstraction procedure and, later, the two stimulation techniques.

\begin{figure}[htbp]
  \begin{center}
    \scalebox{0.9}{\includegraphics[width=\textwidth, trim={0.0cm 0.20cm 0.0cm 0.0cm},clip]{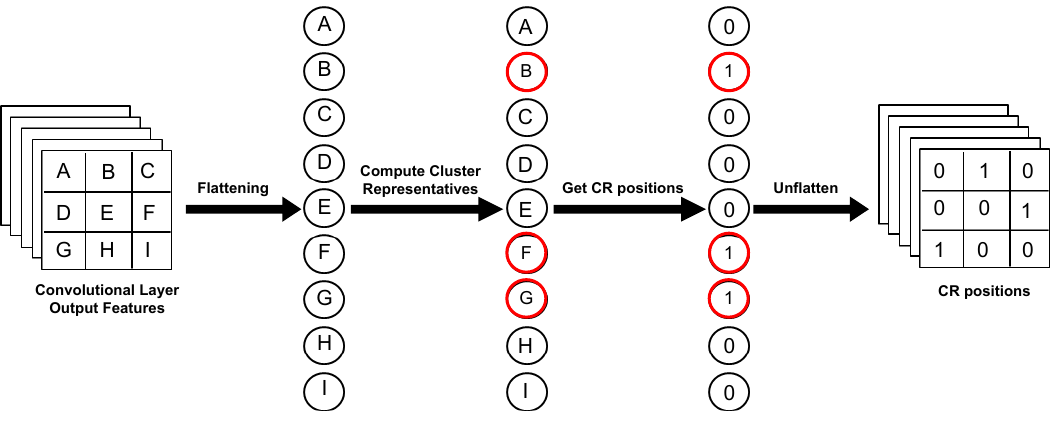}}
  \end{center}
\caption{Obtaining CR positions for one of the features in a convolutional layer. The red mark represent the CR}
\label{Fig: CNN Flatening}
\end{figure}

\vspace{-0.40cm}
\subsection{Identifying Cluster Representatives}
Algorithm~\ref{Alg: Stimulation} depicts our approach for identifying CR.
The inputs to the algorithm are the trained DNN and the dataset.
In line~\ref{Alg 1: method select}, we choose whether to use abstraction.
This selection is done automatically in \toolname which we discuss in Section~\ref{Sec: Tool Architecture}.
For stimulation on the abstract network, we utilise DeepAbstract.
DeepAbstract, as explained in Section~\ref{Sec: Abstraction}, clusters neurons in each hidden layer via the K-means approach and then systematically removes neurons by abstracting them into the CR.
One limitation of DeepAbstract is that it only supports FC layers while image classification models are built based on \emph{convolutional neural network} CNN.
Therefore, we provide a way to work with convolutional neural networks by converting all CNN layers to FC layers, as depicted in line~\ref{Alg: Convert to FC}.
This is possible and indeed a straightforward transformation because a convolutional layer is just a sparsely connected FC layer\cite{ma2017equivalence}.
Nevertheless, our tool is agnostic to the abstraction technique and therefore, DeepAbstract can easily be replaced with a different method that supports other complex layers.
The next step, as in lines~\ref{Alg: deepabstract loop start} $-$ \ref{Alg: deepabstract loop end}, is to run the abstraction function from DeepAbstract.
We then return the abstract network and end the algorithm.

However, abstracting large models with CNN layers would be difficult due to their multidimensional structure.
FC layers are one-dimensional and can directly be clustered, whereas the outputs from CNN layers are two-dimensional features.
For very large networks, converting FC to CNN layers would drastically increase the number of connections, making it difficult to subsume them into CR during abstraction.
Therefore, we propose \approxmethod in which, instead of abstracting the network, we first obtain the positions of CR, which is just the information of the location of the CR neuron.
And then, we stimulate all the CR neurons using the position of CR. 
All the other neurons are not stimulated.
To obtain CR in line~\ref{Alg: k-means on flatten}, we flatten the multidimensional layer output to one-dimensional.
Figure~\ref{Fig: CNN Flatening} depicts the procedure to get the CR positions.
We compute CR via k-means and then store their positions.
We then unflatten the CR position array to get the shape of the original CNN layer and then use it for stimulation.

\vspace{-0.30cm}
  
\subsection{Stimulation Techniques}
This section explains how to identify CNC via stimulation.
We follow a similar approach as in ABS~\cite{liu2019abs}.
The difference is that ABS stimulates all the neurons in the network sequentially, whereas we only stimulate the CR.
Doing so is straightforward for stimulation over the abstract network because it contains only CR; therefore, we directly run the stimulation function on neurons in the abstract network.
Our intuition is that since our clustering is based on the activation values obtained from the benign dataset, the neurons activated in the presence of a trigger called \emph{trojan neurons} may have lower values on benign images.
However, these trojan neurons would have similar activation values, and they would belong to one cluster.
We analysed a trojan model by comparing activation values of benign and trojan images and found out that our intuition is, in fact, true.
Therefore, amplifying the value of that trojan neuron would emulate the trigger behaviour.
Note that there will also be benign CR, but these neurons will be filtered out during stimulation as they will not affect the network output.

\vspace{-0.40cm}
\begin{algorithm}[htbp]
  \caption{Computing Cluster Representatives}
  \begin{algorithmic}[1]
      \renewcommand{\algorithmicrequire}{\textbf{Input:}}
      \renewcommand{\algorithmicensure}{\textbf{Output:}}
      \REQUIRE $\mathcal{N}$: Trained DNN,\\
      $X_\mathit{test} = \{x_1, \cdots, x_T \}$: benign test data,\\
      $\mathit{abstract}$: if $\mathit{TRUE}$ would abstract the network, otherwise would reshape the layer outputs,\\
      \ENSURE $\hat{\mathcal{N}}$: Abstract Network, and CR.
      \IF {$\mathit{abstract} = \mathit{TRUE}$} \label{Alg 1: method select}
          \STATE Convert all convolutional layers to FC layers \label{Alg: Convert to FC}
          \FOR {layer in hidden layers} \label{Alg: deepabstract loop start}
              \STATE Obtain clusters and CR.
              \STATE Abstract the layer based on DeepAbstract and identify CR.
          \ENDFOR \label{Alg: deepabstract loop end}
          \RETURN DNN $\hat{\mathcal{N}}$.
      \ELSE 
          \FOR {layer in hidden layers}
              \STATE Flatten the outputs of all neurons in the layer \label{Alg: flatten layer outputs}
              \STATE Perform K-means clustering on the layer and extract CR \label{Alg: k-means on flatten}
          \ENDFOR
          \RETURN CR
      \ENDIF
  \end{algorithmic}
  \label{Alg: Stimulation}
\end{algorithm}
\vspace{-0.55cm}

\begin{figure}
    \begin{center}
      \scalebox{0.9}{\includegraphics[width=\textwidth, trim={0.0cm 0.30cm 0.0cm 0.0cm},clip]{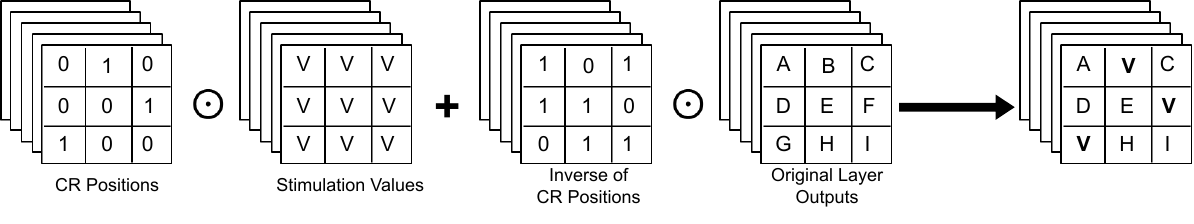}}
    \end{center}
  \caption{\approxmethod method. Here, $\odot$ is the Hadamard product operation. $v$ is the stimulation value}
  \label{Fig: Approximate stimulation}
  \vspace{-0.40cm}
\end{figure}

ABS for CNN layers stimulates each output feature one after the other by setting all the neurons in that feature to the stimulation value (as seen from the source code).
This change can be too strong because all the neurons in one feature can, in reality, never be that high and, therefore, may have a high impact on the network output.
To address this issue, instead of stimulating all the neurons in a feature, we utilise the positions of the CR and stimulate only those neurons.
Figure~\ref{Fig: Approximate stimulation} depicts our idea for stimulation where we apply stimulation values to the CR using the CR position matrix.
This replaces the neuron outputs for only the CR while keeping the neuron values of the rest as it is.
Intuitively, this kind of stimulation would focus only on critical regions in the intermediate feature and, therefore, can produce better results than ABS.
Another advantage is that the stimulation runtime would also decrease because we can skip the features that do not have any CR.
We can call this an approximate method because we stimulate the same neurons as in \abstractmethod.
But unlike \approxmethod, the forward propagation of the stimulation value is on all the neurons including non-CR neurons which may slightly affect the output.

\vspace{-0.20cm}
\section{Tool Architecture}
\label{Sec: Tool Architecture}
\vspace{-0.20cm}

\begin{figure}
    \begin{center}
      \scalebox{0.9}{\includegraphics[width=\textwidth, trim={0.0cm 0.10cm 0.0cm 0.0cm},clip]{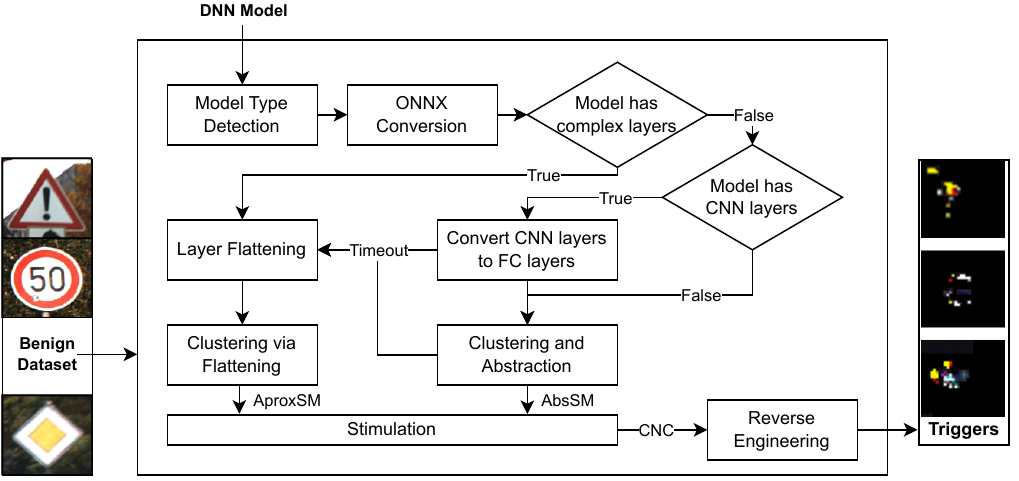}}
    \end{center}
    \vspace{-0.40cm}
  \caption{Tool Framework}
  \label{Fig: Tool Framework}
  \vspace{-0.40cm}
\end{figure}

This section gives an overview of the tool, its functionalities, advantages and limitations. 
Figure~\ref{Fig: Tool Framework} depicts the tool architecture.
\toolname runs on Python and supports state-of-the-art machine learning frameworks such as TensorFlow and PyTorch.
We utilise the \emph{ONNX} library~\cite{onnxlibrary} to convert Pytorch and TensorFlow models to an \emph{onnx file}, which is a generic model that can be loaded into either framework.
Therefore, our tool is compatible with most of the SOTA model architectures.
Additionally, ONNX helps us integrate other abstraction and stimulation techniques with different library and framework requirements.
\vspace{-0.30cm}

The input to our tool is a trained model and a benign test dataset.
The models can be of either TensorFlow or PyTorch frameworks; we automatically convert them to the required framework based on the function to be run.
We have tested the tool's capabilities on several classification model architectures such as \emph{LeNet, AlexNet, VGG, and ResNet}.
More information on these architectures can be found in Sec~\ref{Sec: Experiments}.
We first identify the suitable stimulation method based on the type of model.
Abstract models can be stimulated if the model has only fully connected (FC) and convolutional layers (Conv).
This limitation can, however, be lifted by using a different abstraction technique that supports other layer types.
We switch to the \approxmethod method if the model contains any other complex layers, i.e., layers apart from FC and Conv.
We also set a timeout for stimulation over the abstract network so that we switch to the \approxmethod method if it takes too long to run abstraction.

The output from \toolname are CNCs and the range of stimulation values for compromised neurons at which these neurons trigger the trojan behaviour. 
Next, we use this to run a reverse engineering function that generates masked images that closely represent the trigger.
We utilise the reverse engineering function from ABS. 
However, our tool is agnostic of the type of reverse engineering.
Note that we focus on the traffic sign classification problem in this paper, but the tool is not limited to this application.
The method works on any dataset and applications where a DNN is built.
Finally, we compute the percentage of wrong predictions, i.e., the \emph{Attack Success Rate} (ASR) on the images that contain the reverse-engineered masks.

\vspace{-0.20cm}
\section{Experiments} \label{Sec: Experiments}
\vspace{-0.20cm}
In this section, we show experimental results to validate our method.
Our evaluation is based on the following model architectures:
\begin{itemize}
    \item \emph{FC1:} Flatten, FC($864, 400, 160$), FC($43$),
    \item \emph{FC2:} Flatten, FC($864, 864, 512, 256$), FC($43$),
    \item \emph{FC3:} Flatten, FC($864, 864, 864, 512, 512$), FC($43$),
    \item \emph{NN1:} Conv($8, 16, 32, 16, 8$), Flatten, FC($43$),
    \item \emph{LeNet:} Conv($6$), MaxPool, Conv($16$), MaxPool, Flatten, FC($400, 120$), \\ Dropout($0.5$), FC($80$), Dropout($0.5$), FC($43$),
    \item \emph{AlexNet:} Conv($9$), MaxPool, Conv($32$), MaxPool, Conv($48, 64, 96$), MaxPool, \\ Flatten, FC($864, 400$), Dropout(0.5), FC(160), Dropout(0.5), FC($43$),
    \item \emph{VGG:} Conv($16$), MaxPool, Conv($32$), MaxPool, Conv($64$), MaxPool, Conv($64$), MaxPool, Conv($128$), MaxPool, Conv($64$), MaxPool, FC($1024, 1024$), FC($43$)
\end{itemize}
Here, we represent the type of hidden layer and the respective number of neurons or kernels in the brackets.
For example, FC($864, 400$) means that the network has two fully connected layers, with the number of neurons in the hidden layers being $864$ and $400$, respectively.
The Flatten layer converts a multidimensional layer to one dimension.
Conv denotes convolutional layers, and the number in the bracket represents the number of features/kernels trained in that layer.
MaxPool layers pool neurons by keeping the max value out of the neurons in the pooling window.
A Dropout($0.5$) layer randomly drops out fifty per cent of the neurons.
Notice that all these models end with FC($43$) because the models are built for the dataset with $43$ classes.

We evaluate \abstractmethod with \emph{FC1, FC2, FC3} models and evaluate \approxmethod with \emph{NN1, LeNet, AlexNet, VGG} models.
The reason for this setup is that \emph{FC1, FC2, FC3} are compatible with DeepAbstract, which we use in the \abstractmethod.
We train all the models on the GTSRB dataset\cite{stallkamp2012man} for traffic sign classification.
The GTSRB dataset has over $50,000$ images from German traffic signs that belong to $43$ different classes.
We train all these models on various trigger types as shown in Fig~\ref{Fig: Triggers} where RP, BP, and LRP stand for Red Pixel, Blue Pixel, and Long Red Pixel triggers, respectively.
The numbers $0.2$, $0.4$, and $0.6$ denote the percentage of trigger transparency; a lower number means high transparency.
Note that the trigger is enlarged for visualisation; however, we set the trigger size to be $2.5\%$ of the original image size in the experiments.
We replace $20\%$ of training images with the trigger and change their labels to the target label to prepare the trojan dataset.
Our target class is set to 'stop sign', meaning the output from the trojan model in the presence of a trigger will always output 'stop sign'.
Table~\ref{Table: Model Accuracies} depicts the performance and attack success rates of all these model architectures.

  \newcommand{\width}{1.8cm}
\begin{figure}[t]
  \centering
  \subfigure{
    \includegraphics[width=\width]{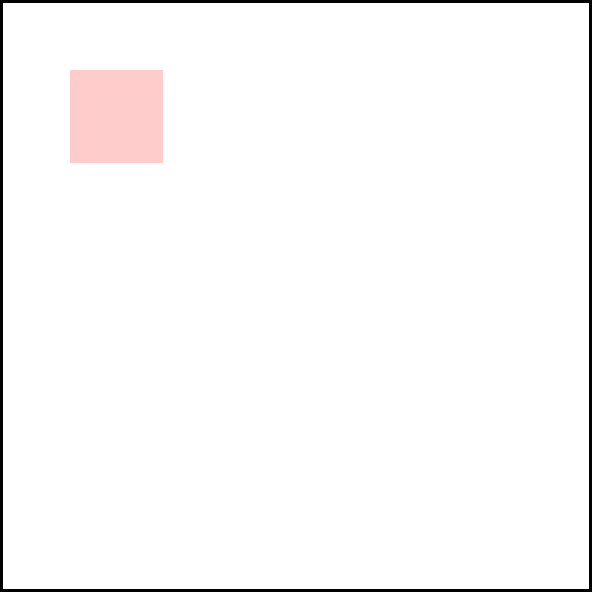}
  }
  \subfigure{
    \includegraphics[width=\width]{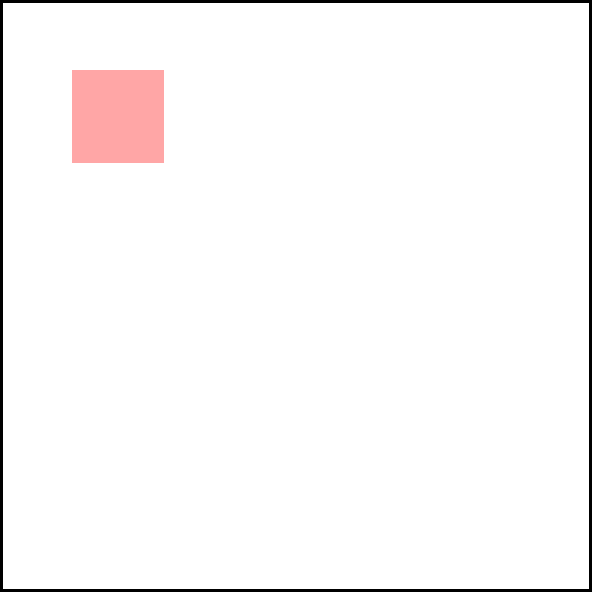}
  }
  \subfigure{
    \includegraphics[width=\width]{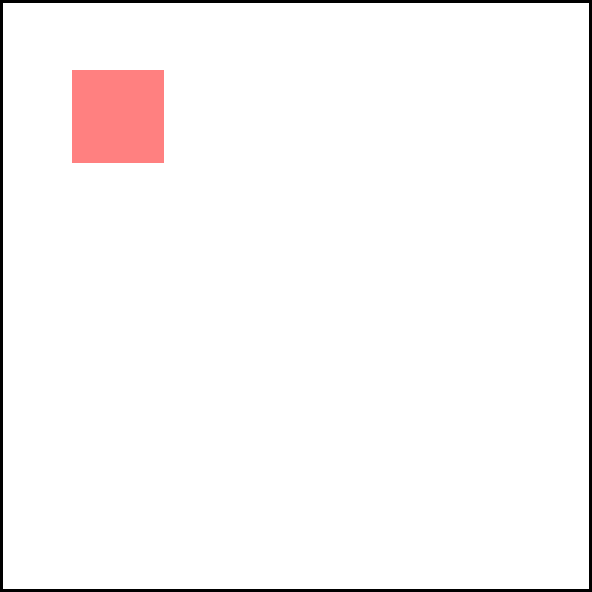}
  }
  \subfigure{
    \includegraphics[width=\width]{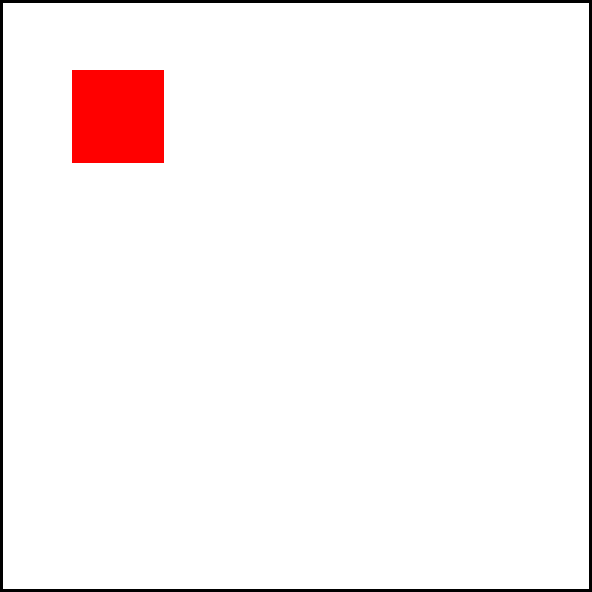}
  }
  \subfigure{
    \includegraphics[width=\width]{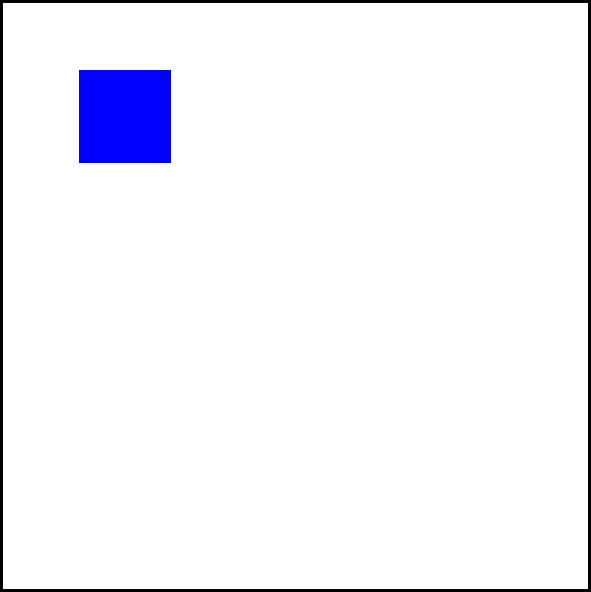}
  }
  \subfigure{
    \includegraphics[width=\width]{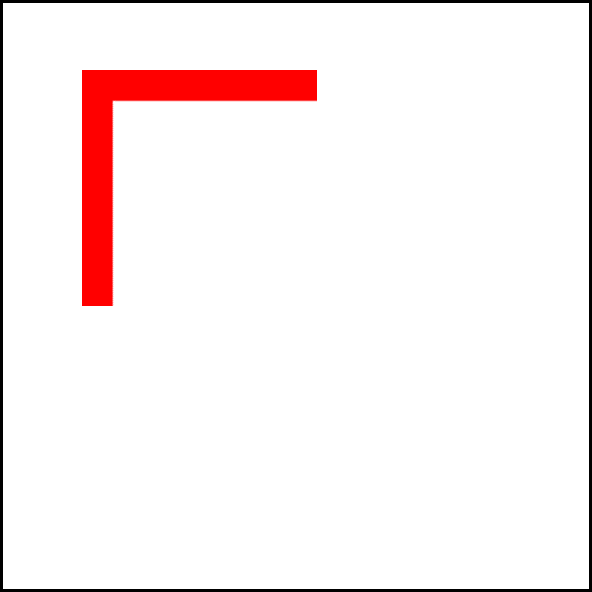}
  }
  \caption{Trojan Triggers. The first four images are Red Pixel (RP) Triggers at transparency $0.2$, $0.4$, $0.6$, $1.0$ respectively. Followed by a Blue Pixel (BP) and a Long Red Pixel (LRP) triggers.}
  \vspace{-0.40cm}
  \label{Fig: Triggers}
\end{figure}

We showcase via experimental evaluations that our methods, \abstractmethod and \approxmethod, perform better in identifying the number of backdoors than other methods.
We evaluate these backdoors by reverse engineering them and computing their ASR.
Both our methods reduce runtime, especially for large models.
We also show that our methods perform better in identifying backdoors even when the trigger is transparent compared to state-of-the-art methods.

\vspace{-0.40cm}
\subsection{Backdoor identification and runtime reduction}
We evaluate our model's performance using three main metrics: number of identified backdoors, ASR, and runtime.
We count the number of identified backdoors by reverse engineering all the CNCs and checking whether the generated trigger can misclassify at least $80\%$ of the classes.
We measure the attack success rate based on the percentage of images with triggers misclassified by the model.
Finally, we measure the runtime, which consists of the time to build the abstraction and identify backdoors.

\vspace{-0.40cm}
\begin{table}[htbp]
  \setlength{\tabcolsep}{4pt} 
  \centering
  \caption{Accuracy and ASR of the trained trojan models}
  \begin{tabular}{llllllll}
      \toprule
        & FC1 & FC2 & FC3 & NN1 & LeNet & AlexNet & VGG\\
      \midrule
      Validation Accuracy & $0.8055$ & $0.8693$ & $0.9371$ & $0,9064$ & $0,9198$ & $0,9506$ & $0.9453$ \\
      Attack Success Rate & $0.9464$ & $0.9997$ & $0.9998$ & $0.9840$ & $0.9792$ & $0.9864$ & $0.9989$ \\
      Trainable Parameters & $3,072$K & $3,987$K & $4,877$K & $80$K & $123$K & $1,264$K & $1,893$K \\
      \bottomrule
  \end{tabular}
\vspace{-0.4cm}
  \label{Table: Model Accuracies}
\end{table}

Table~\ref{Table: Detection Accuracy} depicts the model's performance based on ASR and the number of backdoors.
All the results in the table are averaged over at least $10$ runs for each model architecture, and the trigger used is RP with no transparency.
We separately show the performance of our two methods and compare them to other methods, such as ABS and NC.
For \abstractmethod, we get the abstraction rates $35\%$, $45\%$, $50\%$ for models \emph{FC1, FC2, FC3} respectively by setting the drop in accuracy on test dataset for the abstract model to $5\%$.
On the other hand, the number of CR for the models used in \approxmethod denoted as \emph{Clustering Rate} is set to $10\%$, which means that $90\%$ of the neurons are not simulated.
We see some improvements in detecting the number of backdoors with respect to other methods, but overall, the performance is consistent with the SOTA.
With this setup, although ASR is $100\%$ with our tool and ABS, we see improvements in the number of identified backdoors.

Table~\ref{Table: Detection Accuracy} also depicts the difference in runtime (in sec) with respect to other methods.
We show abstraction and stimulation time separately for our tool.
We see a great reduction in the stimulation runtime for all \abstractmethod models.
This is because the abstract model we use for stimulation is smaller than the original model.
For \approxmethod, the runtime reduction depends on whether cluster representatives are spread out throughout the layer or concentrated at a few features.
If all the CR are concentrated at a few features, then after unflattening the neurons as shown in Fig~\ref{Fig: CNN Flatening}, the rest of the features without CR need not be stimulated.
We do see such behaviour in some models.
This behaviour is because several features that do not propagate large values and do not influence the network output will not become CR.
On the other hand, due to the K-means clustering function being slow, the abstraction runtime is high for models such as \emph{NN1, Lenet, VGG}.
The two factors that increase the K-means clustering time are the number of images ($5000$) in the dataset and the number of neurons in the model.
We can reduce the clustering runtime by reducing the dataset's size or replacing the clustering function.

\vspace{-0.40cm}
\begin{table}[htbp]
  \setlength{\tabcolsep}{4pt} 
  \centering
  \caption{Method Performance}
  \begin{tabular}{clrrrrrrrrrr}
      \toprule
      & \multirow{3}{*}[-0.25em]{Model} & \multicolumn{3}{c}{Attack Succes Rate} & \multicolumn{3}{c}{$\#$ of Backdoors} & \multicolumn{4}{c}{Runtime}\\
      \cmidrule(lr){3-5} \cmidrule(lr){6-8} \cmidrule(lr){9-12} \\
      \addlinespace[-12pt]
      & {} & Ours & ABS & NC & Ours & ABS & NC & \multicolumn{2}{c}{Ours} & ABS & NC \\
      \cmidrule(lr){9-10} \cmidrule(lr){11-11} \cmidrule(lr){11-12} \\
      \addlinespace[-12pt]
      & {} & & & & & & & Abst & Stim & Stim & Stim \\
      \midrule
      \parbox[t]{3mm}{\multirow{3}{*}{\rotatebox[origin=c]{90}{\abstractmethod}}}
      & \emph{FC1}     & $1.0$ & $1.0$ & $1.0$ & $\textbf{10}$ & $7$ & $1$ & $27$ & $\textbf{120}$ & $301$ & $600$ \\
      & \emph{FC2}     & $1.0$ & $1.0$ & $0.94$ & $\textbf{8}$ & $6$ & $1$ & $42$ & $\textbf{131}$ & $175$ & $781$ \\
      & \emph{FC3}     & $1.0$ & $1.0$ & $1.0$ & $\textbf{5}$ & $4$ & $1$ & $63$ & $\textbf{121}$ & $211$ & $906$ \\
      \midrule
      \parbox[t]{3mm}{\multirow{4}{*}{\rotatebox[origin=c]{90}{\approxmethod}}}
      & \emph{NN1}     & $1.0$ & $1.0$ & $0.31$ & $2$ & $2$ & $0$ & $1684$ & $\textbf{156}$ & $174$ & $2,186$ \\
      & \emph{LeNet}   & $1.0$ & $1.0$ & $0.98$ & $12$ & $12$ & $2$ & $99$ & $133$ & $134$ & $632$ \\
      & \emph{AlexNet} & $1.0$ & $1.0$ & $0.07$ & $\textbf{5}$ & $2$ & $0$ & $446$ & $234$ & $236$ & $1,284$ \\
      & \emph{VGG}     & $1.0$ & $0.84$ & $0.84$ & $\textbf{2}$ & $1$ & $1$ & $681$ & $\textbf{330}$ & $342$ & $1,412$ \\
      \bottomrule
  \end{tabular}
\vspace{-0.40cm}
  \label{Table: Detection Accuracy}
\end{table}

Overall, we see improvements in average performance in identifying backdoors while reducing runtime.
This justifies our initial claim that abstracting the network would focus on the important neurons, and stimulating them would lead to identifying more backdoors compared to SOTA.
We have some limitations with runtime due to the slow clustering method. 
However, the function-agnostic architecture of our tool simplifies the limitation as we can use other clustering or abstraction methods.

\vspace{-0.30cm}
\subsection{Performance on various triggers}
Table~\ref{Table: Trojan Model Detection} depicts the method performance concerning the number of backdoors identified when different kinds of triggers are applied.
The numbers in the brackets represent the trigger transparency as explained before in Fig~\ref{Fig: Triggers}.
As we can see, our technique outperforms backdoor identification when the trigger transparency is high.
Since our technique clusters all the neurons with similar values, the neurons representing these transparent triggers could be grouped.
Stimulating this cluster would mean we are enhancing the influence of these neurons while other neuron values remain the same.
Due to this reason, our technique performs better for transparent triggers compared to other methods.

\begin{table}[htbp]
  \setlength{\tabcolsep}{4pt} 
  \centering
  \caption{Number of identified backdoors comparing with ABS}
  \begin{tabular}{clrrrrrrrrrr}
      \toprule
      & \multirow{2}{*}[-0.25em]{Model} & \multicolumn{2}{c}{RP($0.2$)} & \multicolumn{2}{c}{RP($0.4$)} & \multicolumn{2}{c}{RP($0.6$)} & \multicolumn{2}{c}{BP($0.2$)} & \multicolumn{2}{c}{LRP($0.2$)}\\
      \cmidrule(lr){3-4} \cmidrule(lr){5-6} \cmidrule(lr){7-8} \cmidrule(lr){9-10} \cmidrule(lr){11-12} \\
      \addlinespace[-12pt]
      & {} & Ours & ABS & Ours & ABS & Ours & ABS & Ours & ABS & Ours & ABS \\
      \midrule
      \parbox[t]{3mm}{\multirow{3}{*}{\rotatebox[origin=c]{90}{\abstractmethod}}}
      & \emph{FC1} & $\textbf{14}$ & $6$ & $12$ & $12$ & $\textbf{10}$ & $8$ & $7$ & $7$ & $3$ & $4$ \\
      & \emph{FC2} & $\textbf{4}$ & $3$ & $2$ & $4$ & $13$ & $14$ & $2$ & $2$ & $\textbf{9}$ & $8$ \\
      & \emph{FC3} & $\textbf{5}$ & $2$ & $\textbf{11}$ & $4$ & $3$ & $3$ & $\textbf{1}$ & $0$ & $\textbf{6}$ & $5$ \\
      \midrule
      \parbox[t]{3mm}{\multirow{4}{*}{\rotatebox[origin=c]{90}{\approxmethod}}}
      & \emph{NN1} & $\textbf{2}$ & $1$ & $1$ & $1$ & $\textbf{2}$ & $0$ & $0$ & $0$ & $\textbf{2}$ & $0$ \\
      & \emph{LeNet} & $\textbf{7}$ & $5$ & $\textbf{9}$ & $8$ & $\textbf{8}$ & $6$ & $7$ & $7$ & $6$ & $6$ \\
      & \emph{AlexNet} & $\textbf{2}$ & $0$ & $1$ & $1$ & $2$ & $3$ & $\textbf{6}$ & $2$ & $\textbf{5}$ & $3$ \\
      & \emph{VGG} & $0$ & $1$ & $1$ & $1$ & $\textbf{2}$ & $1$ & $\textbf{1}$ & $0$ & $\textbf{1}$ & $0$ \\
      \bottomrule
  \end{tabular}
\vspace{-0.30cm}
  \label{Table: Trojan Model Detection}
\end{table}

Another factor we consider is the influence of the clustering rate on the network's performance.
We do this experiment for only \approxmethod because, for \abstractmethod, we fix the abstraction percentage by setting the drop in accuracy to $5\%$.
Note that the clustering rate would not affect the model performance but instead, the number of CRs obtained.
Table~\ref{Table: Performance at different clustering} depicts the change in runtime and backdoor identification at different clustering rates.
The runtime is linearly dependent on the clustering rate.
The reason is that a lower clustering rate implies a lower number of clusters to be made and in turn, less time required for the computation of the k-means.
On the other hand, the number of backdoors detected varies for each model, and it is important to set the correct clustering rate.
We could improve the way we set the clustering rate by considering different clustering methods such as DBSCAN~\cite{khan2014dbscan}, which does not require the clustering rate to be defined manually but instead optimally finds the best clustering parameter. 
We leave the exploration of other clustering techniques as future work.

All these experiments justify our approach. 
Moreover, developing this as a tool capable of evaluating large SOTA model architectures built on different frameworks makes analysis much easier and systematic.
We want to add more abstraction and stimulation techniques to our tool in the future to improve its performance. 

\vspace{-0.45cm}
\begin{table}[htbp]
  \setlength{\tabcolsep}{4pt} 
  \centering
  \caption{Performance at different clustring rates}
  \begin{tabular}{clrrrrrrrr}
      \toprule
      & \multirow{2}{*}[-0.25em]{Model} & \multicolumn{4}{c}{Clustering Runtime} & \multicolumn{4}{c}{Number of Backdoors} \\
      \cmidrule(lr){3-6} \cmidrule(lr){7-10}  \\
      \addlinespace[-12pt]
      & {} & $10\%$ & $20\%$ & $30\%$ & $40\%$ & $10\%$ & $20\%$ & $30\%$ & $40\%$ \\
      \midrule
      \parbox[t]{3mm}{\multirow{4}{*}{\rotatebox[origin=c]{90}{\approxmethod}}}
      & \emph{NN1}     & $3506$ & $4273$ & $6367$ & $8634$ & $1$ & $1$ & $4$ & $0$ \\
      & \emph{LeNet}   & $106$ & $194$ & $290$ & $389$ & $7$ & $8$ & $5$ & $9$ \\
      & \emph{AlexNet} & $487$ & $1165$ & $1747$ & $2344$ & $2$ & $2$ & $1$ & $1$ \\
      & \emph{VGG}     & $715$ & $1444$ & $2208$ & $3003$ & $0$ & $1$ & $1$ & $0$ \\
      \bottomrule
  \end{tabular}
\vspace{-1.15cm}
  \label{Table: Performance at different clustering}
\end{table}

\section{Conclusion} \label{Sec: Conclusion}
\vspace{-0.20cm}

In this paper, we have introduced the tool \toolname for finding backdoors in DNNs.
We have provided experimental evidence that the two analysis methods implemented are superior in terms of performance to state-of-the-art methods.
The two methods are complementary in that the best choice of the method depends on the model under consideration.

In the future, we will improve our tool in several dimensions further.
As one of the next steps, we will consider different abstraction methods, which will speed up the computation of abstract DNNs to be used for analysis.
We will also plan to further reduce the runtime by using parallel computation approaches.  

\bibliographystyle{splncs04}
\bibliography{Bibliography}

\end{document}